\documentclass[sigconf]{acmart} 

\usepackage{ascmac}
\usepackage{tabularx}
\usepackage[shortlabels]{enumitem}

\AtBeginDocument{%
  \providecommand\BibTeX{{%
    \normalfont B\kern-0.5em{\scshape i\kern-0.25em b}\kern-0.8em\TeX}}}

\copyrightyear{2024}
\acmYear{2024}
\setcopyright{acmlicensed}\acmConference[AHs 2024]{The Augmented Humans International Conference}{April 4--6, 2024}{Melbourne, VIC, Australia}
\acmBooktitle{The Augmented Humans International Conference (AHs 2024), April 4--6, 2024, Melbourne, VIC, Australia}
\acmDOI{10.1145/3652920.3652922}
\acmISBN{979-8-4007-0980-7/24/04}

\begin{document}

\title{FastPerson: Enhancing Video-Based Learning through Video Summarization that Preserves Linguistic and Visual Contexts}

\author{Kazuki Kawamura}
\affiliation{
\institution{The University of Tokyo, Tokyo}
\country{}
}
\affiliation{
\institution{Sony CSL Kyoto, Kyoto}
\country{Japan}
}
\email{kwmr@acm.org}

\author{Jun Rekimoto}
\affiliation{
\institution{The University of Tokyo, Tokyo}
\country{}
}
\affiliation{
\institution{Sony CSL Kyoto, Kyoto}
\country{Japan}
}
\email{rekimoto@acm.org}

\renewcommand{\shorttitle}{FastPerson: Effective Video Summarization Preserving Linguistic and Visual Contexts}

\renewcommand{\shortauthors}{Kawamura et al.}

\begin{abstract}
Quickly understanding lengthy lecture videos is essential for learners with limited time and interest in various topics to improve their learning efficiency.
To this end, video summarization has been actively researched to enable users to view only important scenes from a video.
However, these studies focus on either the visual or audio information of a video and extract important segments in the video.
Therefore, there is a risk of missing important information when both the teacher's speech and visual information on the blackboard or slides are important, such as in a lecture video.
To tackle this issue, we propose FastPerson, a video summarization approach that considers both the visual and auditory information in lecture videos. 
FastPerson creates summary videos by utilizing audio transcriptions along with on-screen images and text, minimizing the risk of overlooking crucial information for learners.
Further, it provides a feature that allows learners to switch between the summary and original videos for each chapter of the video, enabling them to adjust the pace of learning based on their interests and level of understanding.
We conducted an evaluation with 40 participants to assess the effectiveness of our method and confirmed that it reduced viewing time by 53\% at the same level of comprehension as that when using traditional video playback methods.
\end{abstract}




\maketitle

\section{INTRODUCTION}
The proliferation of video-sharing platforms has significantly enhanced the use of videos across various fields, including entertainment (film), music, blogging, and notably, education.
As an example of using videos in education, flipped learning inverts the conventional classroom model by encouraging students to engage with video lectures and content before class. 
This preparatory engagement enables in-class sessions to be dedicated to more interactive discussions, problem-solving activities, and hands-on learning experiences, enhancing the educational process~\cite{Betihavas2016}.
Further, the rise of massive open online courses (MOOCs) has democratized access to education, enabling a global audience to participate in high-quality learning experiences. 
MOOCs offer an extensive array of video lectures and educational resources, enabling learners worldwide to access and engage with course materials at their own pace~\cite{Hew2014, Philip2014}.

Many previous studies have confirmed the effectiveness of these video-based learning methods. 
For example, Zhang et al. showed that interactive videos improve learners' comprehension~\cite{Zhang2006}.
Further, Kay et al. underscored the efficacy of video-based learning (VBL), showing its positive impact on comprehension, adaptation, and achievement~\cite{Kay2012}.
Although the results of these studies illustrate the benefits of video-based learning methods, challenges such as the time-consuming search for appropriate videos and difficulty in adjusting the learning pace to individual comprehension and interests continue to persist.
In fact, Chen et al. showed that users spent 80\% of their time on video-sharing platforms for browsing and partial viewing~\cite{Chen2013}, which suggests that efficiently finding the desired content from a large number of videos is a major challenge faced by current video-sharing platforms.
In addition, online learning platforms such as YouTube's education channel, Coursera, and EdX contain videos that are 30 min to 2 h long. 
However, unlike written content, videos make it difficult to obtain an overview quickly by skimming or adjust the amount of information viewed for each chapter.
This limitation is a barrier to learners using videos as a learning tool.

\begin{figure*}[t]
  \centering
  \includegraphics[width=0.98\linewidth]{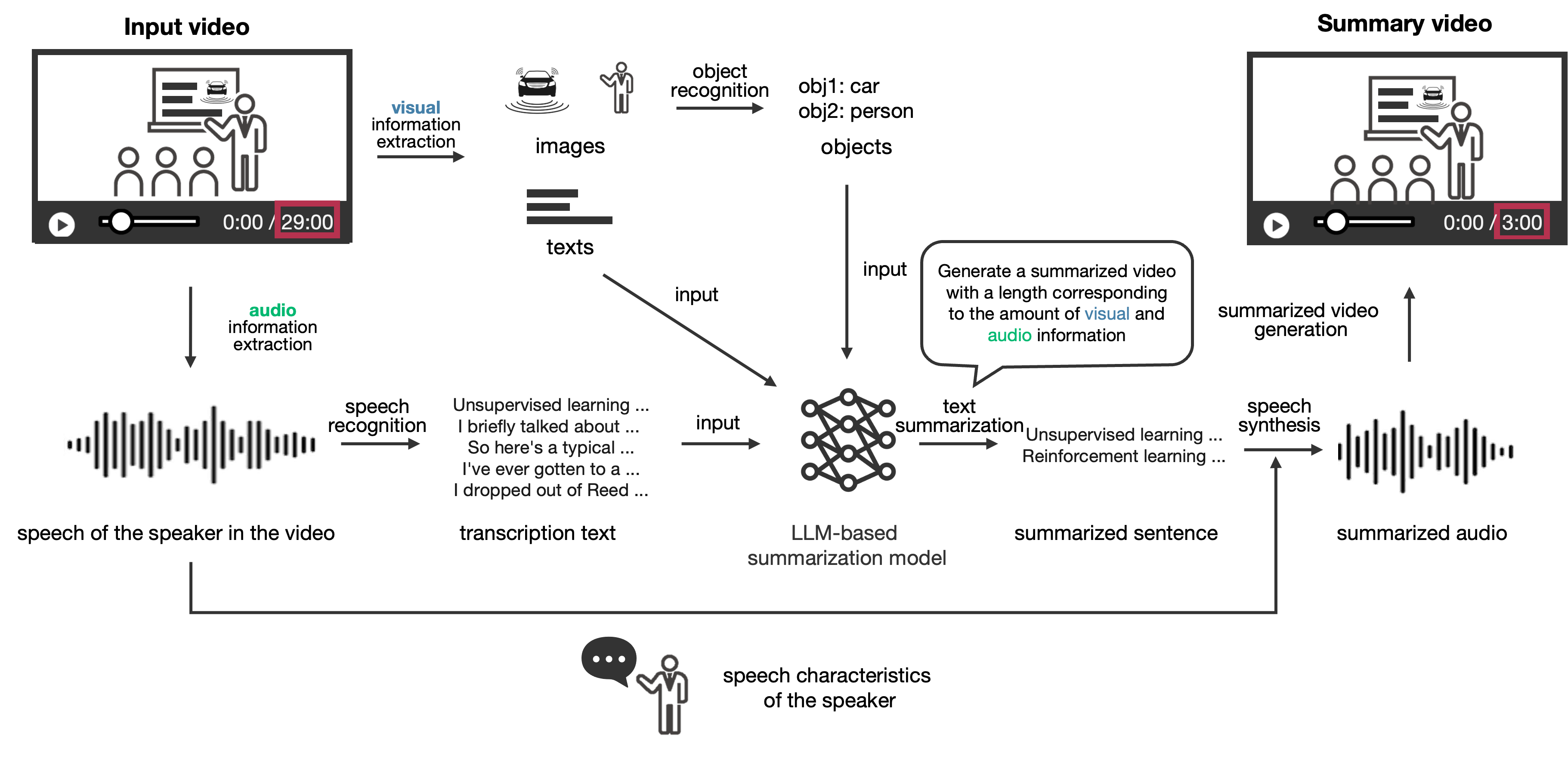}
  \caption{FastPerson: Video summarization method that generates a summary sentence using visual (objects and sentences on the frame) and audio information (speaker's voice) before synthesizing the sentence into speech using the speech characteristics of the original speaker.}
  \label{fig:architecture}
\end{figure*}

To address these problems, video summarization methods have been proposed for helping users grasp the content of videos in a short period of time, and various methods for extracting key information have been studied~\cite{Tu2007, Money2008}.
These existing methods focus on visual or audio information to determine the important parts of the video and generate a summarized video by joining the important parts.
However, these approaches are limited in lecture videos, because in narrated videos, interesting or useful elements are present on the screen and in the audio, and they may not always overlap.
For example, in an educational video where the instructor points to a specific diagram and provides an explanation, the importance of the explanation is conveyed by the audio, while the diagram appears in the video. 
If the viewer listens only to the audio, he or she may miss the visual explanation and not fully understand the importance and content of the diagram.

To overcome these limitations, we propose a new video summarization technique referred to as FastPerson, which comprehensively considers both visual and audio information.
In contrast to conventional video summarization, which extracts important parts of a video, the proposed method creates a summary text that considers both visual and audio information before generating a summary video using narration of the text.
In addition, the summary rate is adjusted to reflect the amount of visual and audio information in the summary video to avoid presenting over- or under-information while preserving the important content for the learner. 
This results in more detailed summaries for the content-rich parts of the video, such as when the speaker speaks for a long time or when slides with a lot of textual information are shown. 
This reduces the risk of learners missing important visual or audio information.

The present paper also focuses on the design of interaction methods to provide a user-centered learning experience, addressing the challenge that summary videos do not allow adjusting the amount of information viewed on a chapter-by-chapter basis to enable users quickly grasp the outline of the content. 
To overcome this limitation, the proposed system provides summary and original videos for each video chapter, thereby allowing the user to switch between them by pressing buttons. 
This allows learners to choose which video to watch based on the content, interest, or comprehension. 
For example, the summary video allows the learner to obtain an overview of each chapter quickly, and if more detailed information is required, the learner can watch the full original video of that chapter. 
Indeed, it is possible to only watch the summary video to obtain an overview of the video. 
Further, the title, summary text, and thumbnails of each video chapter are displayed in a list.
The learner can directly select and watch only chapters that interest them.

We conducted an evaluation experiment with 40 participants to assess the learning effectiveness and usability of the proposed method. 
After watching two lecture videos, we administered a quiz on the videos and confirmed that using FastPerson reduced the viewing time for the two videos by 53.24\% and 53.11\% and the users could understand the videos at the same level as that when they watched the original videos. 
In addition, although 78\% of the users rated the ability to switch between the summary and original videos as useful, the results suggest that several improvements can be made, including improvements to the transition between video chapters and the audio quality of the summary video. 

The main contributions of the current paper are as follows:
\noindent
\begin{itemize}
    \item We propose a method, "FastPerson," which generates summary videos reflecting important information from both the visual and audio elements of lecture videos by comprehensively considering them.
    \item We suggest a user-centered learning experience design that enables users to switch between the original and summary videos, allowing them to choose between concise and detailed information acquisition in each chapter of the video.
    \item We demonstrate that applying our proposed method to video-based learning can enable learners to efficiently acquire information from lengthy video content.
\end{itemize}

\section{Related Work}
\subsection{Video-Based Learning}
In this contemporary digital era, VBL has become pivotal for education.
This transformation can be attributed to the widespread availability of the internet and the proliferation of online learning platforms.
MOOCs, exemplified by platforms such as Coursera and EdX, have played a pivotal role in steering this shift toward video-centric pedagogy.
For example, studies by Khe et al. and Guo et al. demonstrated the importance and expanded the use of video lectures in MOOCs ~\cite{Hew2014, Philip2014}.
Furthermore, video tutorials featured on platforms such as Khan Academy and Udemy are instrumental in democratizing skill acquisition, spanning a diverse spectrum from coding to the culinary arts~\cite{Clive2011, Iuliana2018}.
In professional settings, platforms such as LinkedIn Learning have harnessed the power of video content to facilitate career development and training.
These examples suggest that video-based resources have become omnipresent in modern education and training and are widely used to acquire knowledge and skills.
The instructional videos often involve an instructor speaking and using slides to provide explanations.
Although there are a small number of videos that convey information and skills in VBL without relying on sound, such as silent demonstration videos, we focus on videos that are more commonly rich in audio and visual information.

The efficacy of video as a pedagogical tool has been extensively researched and validated~\cite{sablic2021}.
Zhang et al. found that interactive videos, which allow learners to engage with content, enhance comprehension~\cite{Zhang2006}.
Kay's research on video podcasts revealed their positive influence on students' understanding and achievement~\cite{Kay2012}.
Although traditional teacher-centered teaching styles often leave students reluctant and unable to fully develop critical thinking skills, instructional video enable a shift to learner-centered learning~\cite{akccayir2018}.
Masats and Dooly's research showed that the use of video for instructional purposes introduced innovative and creative teaching perspectives~\cite{masats2011}.
Furthermore, Brecht et al. claimed that learners who could not fully understand the instructional content in lectures or textbooks could learn by repeatedly viewing the lessons as required to meet their individual learning needs~\cite{brecht2008}. 
However, it remains difficult to control which parts of a video are important to view for their needs simply by skipping or adjusting the playback speed.
Our system addresses this challenge by providing an active, personalized learning experience.

\subsection{Video Summarization}
Video summarization involves creating short-form content from an original longer video so that users can quickly grasp the main content of interest. 
Video summarization methods can be divided into three categories:
\vspace*{2mm}
\begin{itemize}
  \item Methods based on visually important elements
  \item Methods based on aurally important elements
  \item Methods based on elements of importance to the learner
\end{itemize}
\vspace*{2mm}

\textbf{Visual-element-based summarization methods} have been continuously studied in the field of computer vision to identify important frames or segments using visual information. 
In early work, supervised learning methods predicted the importance of each segment face of a video from video features such as faces, objects, and aesthetic features~\cite{gygli2014}. 
Recent methods incorporate high-level features extracted from deep neural networks such as convolutional neural networks (CNN)~\cite{fukushima1980, lecun1989} and recurrent neural networks (RNN)~\cite{hochreiter1997} to improve video summarization performance~\cite{gygli2014, yao2016, zhang2016, zhao2017}.
Attention mechanism~\cite{zhou2018}, graph convolutional network~\cite{park2020}, and Transformer~\cite{liu2020} have also contributed to improving the performance of video summarization. 
Other methods detect video segments that have objects related to the title~\cite{Song2015}. 
Although these methods are promising for video surveillance and sports video highlight generation, their application to videos that display slides or boards with considerable textual information on the frame, such as online lectures and how-to videos, which are the target of the present study, is yet to be explored. 
Further, there are several methods to predict important segments of a video using audio features in addition to visual features, especially for cooking and sports videos~\cite{shang2021, zhao2021}. 
However, it is very difficult to prepare such data for lecture videos covering a wide range of fields, and therefore, we aim to establish a technique that can generate summarized videos for lecture videos without supervised data by using a large language model (LLM).

\textbf{Auditory-element-based summarization methods} convert the audio in a video into transcribed text and apply document summarization techniques to this text for generating a summary text of the video~\cite{chopra2016, rush2015}.
The summarization technique~\cite{palaskar2019}, which uses the Seq2seq model, is an example of such a technique and can generate a summary while modifying the content of the original text.
In recent years, there has been an increase in the number of examples that use LLMs, such as ChatGPT~\cite{lund2023}, to generate video summary documents\footnote{\url{https://glasp.co/youtube-summary}}.
These methods consider the importance of linguistic information and can effectively understand the content of audio-based videos such as lectures and presentations.
However, the information provided by these document-centered summarization methods neglects the importance of visual elements in videos. 
In lecture videos, visual information such as slides and charts play an important role in many scenarios, and it is difficult for learners to understand the content of a video by reading only the summary text of the speaker. 
Recognizing the importance of the speaker's speech in a video and believing that the visual representation of the video is also important, we aim to establish a video summarization technology that considers both audio and language information and produces video rather than text as the output. 
There are other methods that focus on the pitch rather than the content and identify high-pitched parts as important segments of a lecture video~\cite{he1999}; however, they provide less information than the speech content and are less robust to general noise. 
Therefore, our method uses the speech content directly for summarization. 

\textbf{Learner-preference-based methods} include many proposed video summarization methods that consider interaction history and user preference information~\cite{agnihotri2005, del2017}.
There are methods for generating summaries based on user interaction data, such as Lecturescape~\cite{kim2014}, which retrieves frames frequently viewed by other learners, and EpicPlay, which identifies key moments in sports videos by leveraging social media viewer interaction information~\cite{Anthony2012}.
Video summarization based on the interest level of individual users has also been developed. 
For example, Varini et al. proposed a method for generating personalized travel summary videos by users directly providing their preferred genres~\cite{varini2015}.
EgoScanning allows users to enter keywords related to the events they are interested in to speed up the playback of the non-corresponding parts~\cite{higuchi2017}. 
Further, ElasticPlay has a feature that allows users to interactively adjust the overall length of the summary video~\cite{jin2017}.
Our method does not directly input or estimate user preferences, and instead, it allows each user to freely adjust the pace of video viewing according to their own preferences.

\section{FastPerson's Video Summarization Method}
FastPerson can create a summarized video that considers both visual (text and objects on the screen) and audio information (speaker's voice).
FastPerson automatically divides the original video into multiple coherent video segments, providing the user with the summarized video and original video for each segment.
The quality of the summary and original videos are matched by combining this summary text with a synthesized
voice of the same voice quality as the speaker of the original video and the corresponding video of the original video, thereby providing the user with a seamless transition between the summary and original videos. 
In addition, when generating the summary video, the summary ratio is adjusted by reflecting the amount of visual and audio information to avoid over- or under-information while preserving important content for the learner. 
This method is expected to be applied to a wide range of lecture videos, and therefore, the use of LLM-based summarization is designed to eliminate the need for training with supervised data. 
We describe the summarization method by focusing on the FastPerson architecture, as shown in Fig.~\ref{fig:architecture}.

\subsection{Video Chapter Segmentation} 
FastPerson first detects scene transitions and silence in a video and divides them into segments to summarize the video into coherent chapters and generate a video summary. 
For scene transition detection, we use a technique proposed by Nisreen et al.~\cite{Nisreen2012} to analyze the variation of the color distribution (histogram) between consecutive frames and determine the boundaries of the segments based on the magnitude of the variation. 
According to Truong et al.~\cite{Tu2007}, distinct variations between consecutive frames indicate a change in scene, and histogram variations indicate changes in the mood or background of the video. 
We use an algorithm that detects periods when the sound waveform has an amplitude below a certain threshold as silence because visual and audio breaks do not always coincide. 
This method is based on the work of Scheirer et al., which enables detecting periods of low amplitude or the absence of certain frequencies while identifying segments of silence or stillness~\cite{Scheirer1997}. 
These visual scene transitions and periods of silence in audio are considered to separate and segment the video at the cut point. 
This functionality enables automatic chaptering (segmentation) without the need for the video creator to tag each contiguous segment.

\subsection{Visual and Audio Information Extraction}
Our method converts visual and auditory information into textual information for generating a summary video that considers both. 
We use optical character recognition (OCR) and object detection techniques to obtain visual information. 
OCR is used for digitizing imaged document information, and it can detect characters in images and videos and convert them into documents~\cite{Smith07}. 
This technology can effectively capture document information in video images, such as the content of slides or handwritten notes on a whiteboard in a lecture video. 
This information can then be used to reflect the topic and content of the documents in the video in the summary.
Object detection is a technique for detecting and identifying specific objects or entities in each frame of a video~\cite{Cortes15}.
The objects in the video are often closely related to the story focusor theme of the segment.
If a car is shown on a slide in a scene from an online lecture on machine learning, the speaker is expected to talk about automated driving or image recognition, and the object ``car'' is considered a key point in that video segment. 
Therefore, key scenes and important entities in the video can be identified using object detection, and effective summaries can be generated based on this information. 
Our system uses the Tesseract OCR engine~\cite{Smith07} for OCR and the faster R-CNN~\cite{Cortes15} with ResNet-50-FPN as the object model backbone. 

The accurate extraction of speech information is necessary for academic content in videos, especially lectures and seminars. 
Audio information is extracted from a video as follows:
\begin{enumerate}[(1)]
  \item An audio stream is extracted from the video file.
  \item This audio stream is fed as an input to a speech recognition model to obtain a transcription of the document. 
\end{enumerate}
The speech signal needs to be converted into document data with high accuracy because the accuracy at this stage can affect the quality of the summary.
Recent advances in speech recognition have been attributed to models based on RNN~\cite{Graves2013, hannun2014} and Transformer~\cite{Vaswani2017, wang2020}. 
In our method, one of these models, namely Whisper, is used as the speech recognition model~\cite{radford2023}. 
Whisper has been trained on approximately 680,000 h of multilingual speech collected from the Web, and its recognition accuracy has been shown to be comparable to that of humans. Despite the high performance of this method, the recognition rate is not zero. 
Therefore, we use OCR to recognize the text in the video to cover the misrecognition in speech and ensure that important words in the video are not omitted.

\subsection{Summary Audio Generation}
Video summarization uses data labeled with important points in a video, along with supervised learning methods to detect important locations in the video. However, our target educational videos cover a wide variety of domains, and it is difficult to use training data to extract such important points.
Recently, a number of LLMs have been proposed in the natural language processing domain based on Transformer architectures, including BERT~\cite{devlin2018}, T5~\cite{raffel2020}, and GPT~\cite{radford2018, radford2019, brown2020}. 
These models are trained on documents and dialogues from different domains and have a wide range of domain knowledge, hence enabling them to perform various natural language processing, including summary generation~\cite{basyal2023}. 
Our system can summarize lecture videos without the need for teacher data using this LLM.  
Usually, when a summary is generated from speech, a transcript of the speech is used as input to generate the summary. 
However, this method generates a summary based on both the transcript of the speech and visual metadata obtained through OCR and object recognition, thereby reflecting the visual information of the video in the summary. 
The following instructions are entered into LLM~\cite{ray2023} to generate a summary that includes both visual and audio information, and then, a document summary is generated for each video segment.

\hspace{3mm}
\begin{itembox}[l]{Instructions}
  Using the provided transcription of spoken content, OCR-derived textual data, and object detection information, synthesize a comprehensive summary. This summary should highlight the key themes and actions depicted in the video. Focus on distilling the essence of the video by combining insights from the transcription (which captures the spoken words), the OCR data (which provides text found within the video), and object detection (which identifies significant objects and actions). 
\end{itembox}
\hspace{3mm}

Along with these instructions, the transcribed text of each video segment, OCR, and resulting visual information from object recognition are entered into the LLM as documents.
This method enables creating a summary that emphasizes nuances not included in the speech content, speaker's movements, visual objects, and documents within slides and visual presentations.

\subsection{Summary Length Calculation}
The length of the document summary must appropriately reflect the amount of information in the original video when summarizing a video.
In other words, if the video is long or rich in content, it is necessary to provide sufficient information in the summary, and if the original video is short, the summary should be shortened by the same amount to eliminate redundancy.
Therefore, we adjust the number of characters in the output summary based on the length of the transcribed document.
In addition, the number of output characters after summarization is adjusted based on the amount of visual information because our method generates video summaries based on visual information as well as audio.
For example, if the amount of text on a slide is large even if the speaker's speech is short, it must be summarized to accurately reflect the information content of the slide.
From this point of view, the proposed system calculates the number of output characters \(N \) of the summary using
\begin{align*}
  N = \max(50, w_s * L_t + w_i * (L_o + L_c)),
\end{align*}
where
\begin{itemize}
    \item \( N \) represents the final output word count of the summary.
    \item \( L_t \) is shown in the length of the transcribed text.
    \item \( L_o \) represents the number of objects identified on the screen.
    \item \( L_c \) represents the word count of OCR-processed visual elements.
    \item \( w_s \) and \( w_i \) represent the weight coefficients to adjust the importance between audio and visual information.
\end{itemize}
The weight coefficients \( w_s \) and \( w_i \) are selected based on experimental data and user preferences. 
These coefficients can be varied to adjust the emphasis on visual or auditory information. 
Although the balance can be altered according to user interest and comprehension levels to provide personalized summary videos, in the present study, fixed values of 0.3 and 2.5 are set for \( w_s \) and \( w_i \), respectively.

\subsection{Summary Video Generation}
A conventional audio summarization method used for audiobooks evaluates the importance of each audio transcription on a sentence-by-sentence basis and directly extracts and combines important audio segments. 
However, this method poses a challenge to the continuity of audio and video when merging the extracted segments. 
This method cannot produce a video summary that reflects the visual representation of the video. 
Therefore, this method uses a speech synthesis~\cite{Shen2018, Vandenoord16} method, which uses a document summary with visual and audio information as the input and speech synthesis technology to generate its reading voice.
After the synthesized speech is generated, the optimal video segment is selected from the original video based on the length of the speech. 
The video segment is cut from the beginning, middle, or end of the original video, depending on the length of the summarized video. 
Based on the results of a simple user experiment, the video segment in the middle of the original video is used as the default joining method. 
A continuous and concise video summary is achieved by combining this selected video segment with the synthesized audio. 
This method avoids unnatural connections caused by combining different parts of the video. 
In addition, users are expected to frequently switch between the original and synthesized videos when using this method, and therefore, it is necessary to provide a seamless experience. 
Thus, we adopt a method that minimizes the change in the audio between the original and summarized videos using the speaker adaptation technique when synthesizing the audio~\cite{Arik2017}. 
The features of the speaker's voice in the original video are extracted, and a synthesized voice is generated that reflects the speaker's voice quality. 
This method reduces the mismatch of voice features when switching between the original and synthesized videos, thereby enabling seamless switching.

\begin{figure*}[t]
  \centering
  \includegraphics[width=\linewidth]{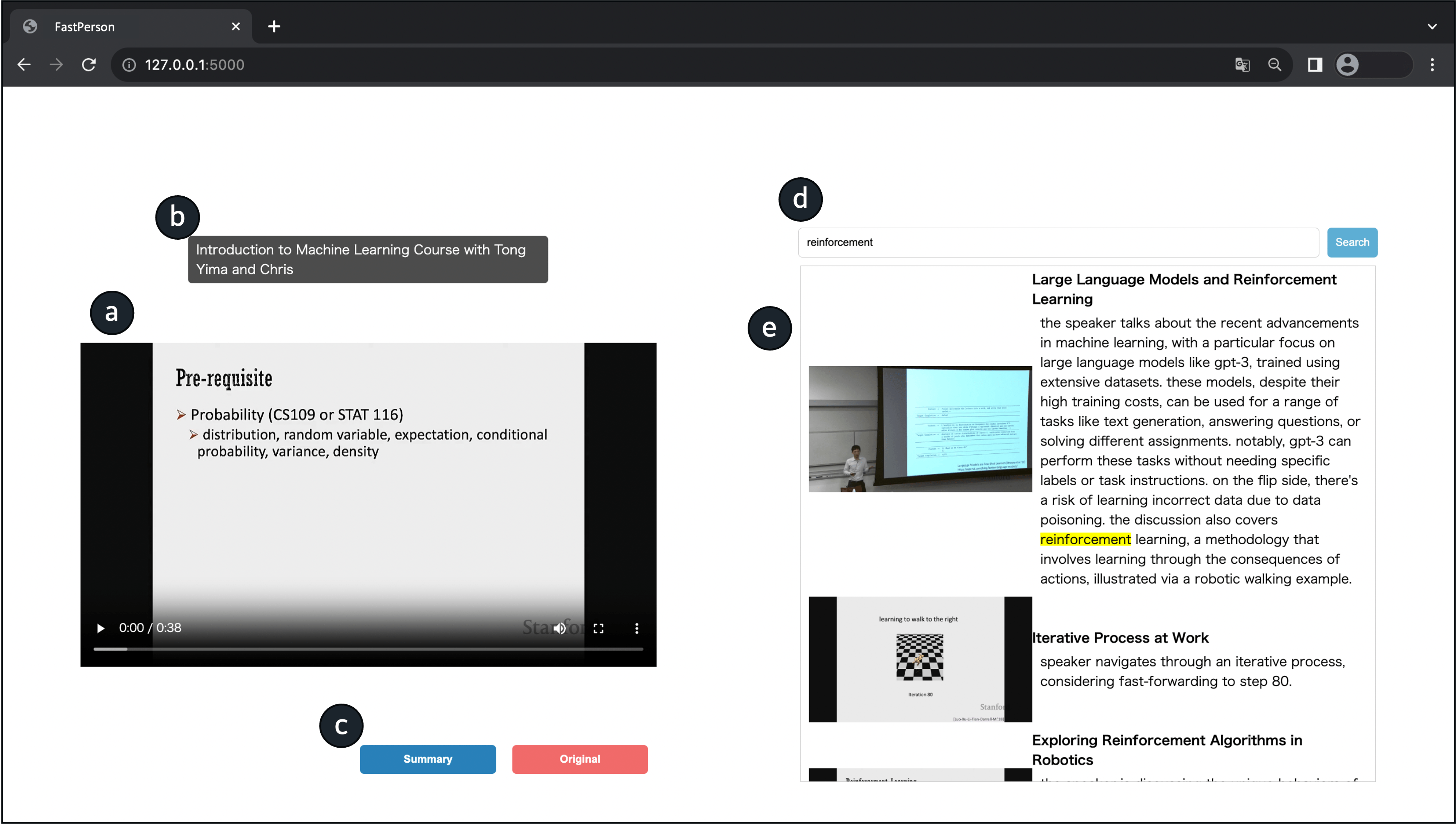}
  \caption{Overview of the FastPerson user interface. \textbf{A. Video Window} showcases the current video playback and offers user-controlled speed adjustments. \textbf{B. Video Title} provides segment-specific titles. \textbf{C. Summary and Original buttons} allow seamless toggling between video versions and facilitate segment switches. \textbf{D. Search Container} offers a text-based search mechanism. \textbf{E. Segments Container} displays the thumbnails of video segments, ensuring quick navigation.}
  \label{fig:interface}
\end{figure*}

\section{FastPerson application and interaction}
\subsection{Application}
FastPerson enables users to switch between the original longer video and summarized video for each segment within each video, depending on their level of interest and understanding. 
In addition, a static component placed on the side of the video displays a thumbnail image and summary text for each video segment, enabling the user to quickly access the corresponding video chapter.

The entire user interface is shown in Fig.~\ref{fig:interface}; \textcircled{\scriptsize a} \textbf{Video Window} allows users to view videos in the window in either a summarized or original format.
The playback speed can be adjusted based on preference, enabling flexible viewing of content.
\textcircled{\scriptsize b} \textbf{Video Title} is placed at the top of the video window and displays the title of the segment that is currently playing. 
\textcircled{\scriptsize c} \textbf{Summary and Original buttons} allows the user to quickly switch between the summary and original full version of the video.
In this case, making the voice quality of the summary video similar to that of the original video allows the user to easily perceive continuity and maintain the consistency of the video experience.
Next to the video is a window displaying a pair of text summaries and thumbnail images representing the chapters of the video. 
These windows are effective in helping users quickly access the segment of interest in the video~\cite{chi2012, pavel2014}. 
\textcircled{\scriptsize d} \textbf{Search Window} provides a document entry box and a search button to search for specific keywords in a video segment. 
This feature allows the user to see where their topics of interest are discussed in the video. 
\textcircled{\scriptsize e} \textbf{Segment Window} displays thumbnails, titles, and summary text for each video segment. 
Users can quickly access the appropriate video segment by clicking on the appropriate thumbnail.

The application is built using Flask, a lightweight Python web framework.
The FastPerson backend manages the files for each video segment and delivers them efficiently.
The backend includes functions to load the associated thumbnail and text files based on the name of a specific segment in the video, enabling a quick delivery of the video segment requested by the user.
On the front end, the user interface is built using HTML, CSS, and JavaScript.
The interface includes a video player, video title display, a button to toggle between the summarized and original videos, a search function, and thumbnails and summaries for each segment.
JavaScript is used to control video playback, implement the search function, and respond to user interactions.

\begin{table*}[t]
  \centering
  \caption{Details about the videos used in the evaluation experiment}
  \begin{tabularx}{\textwidth}{@{}l *{2}{>{\raggedright\arraybackslash}X}@{}}
  \toprule
  & \textbf{Video 1} & \textbf{Video 2} \\
  \midrule
  Title & Origin of Life & Market Revolution \\
  Thumbnail & \raisebox{-0.5\totalheight}{\includegraphics[width=5cm, height=2.5cm, keepaspectratio]{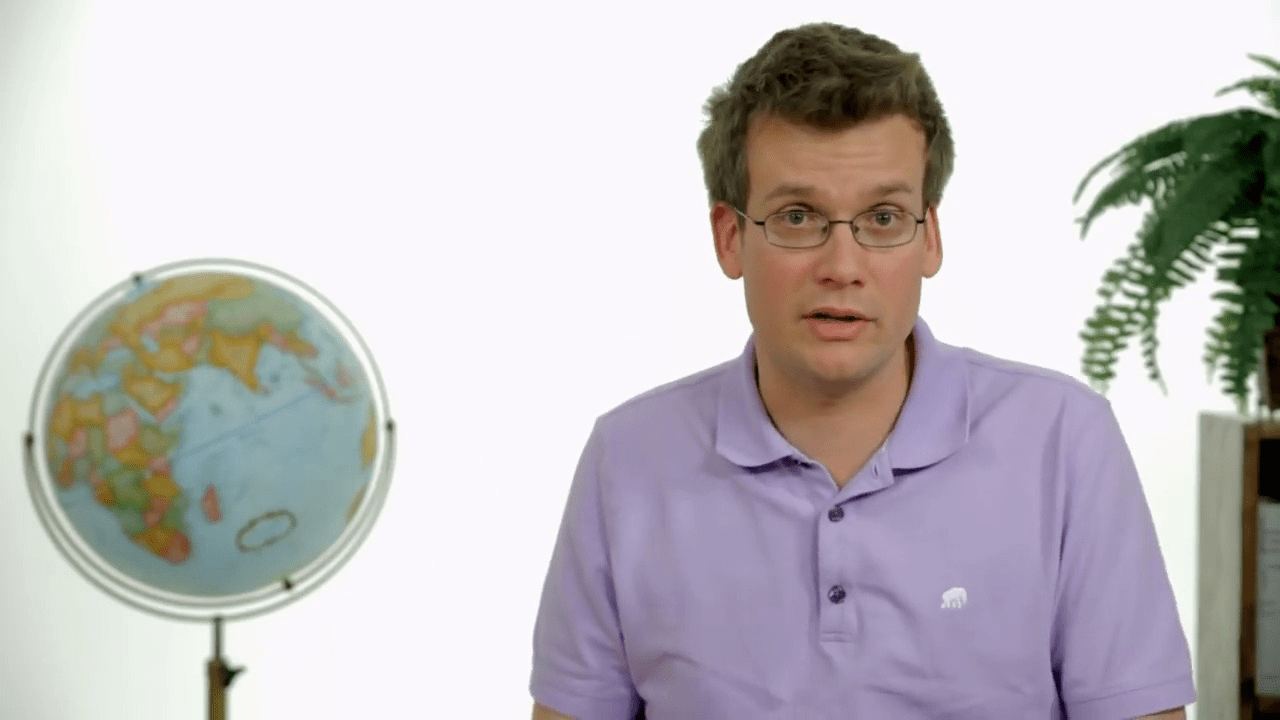}} & \raisebox{-0.5\height}{\includegraphics[width=5cm, height=2.5cm, keepaspectratio]{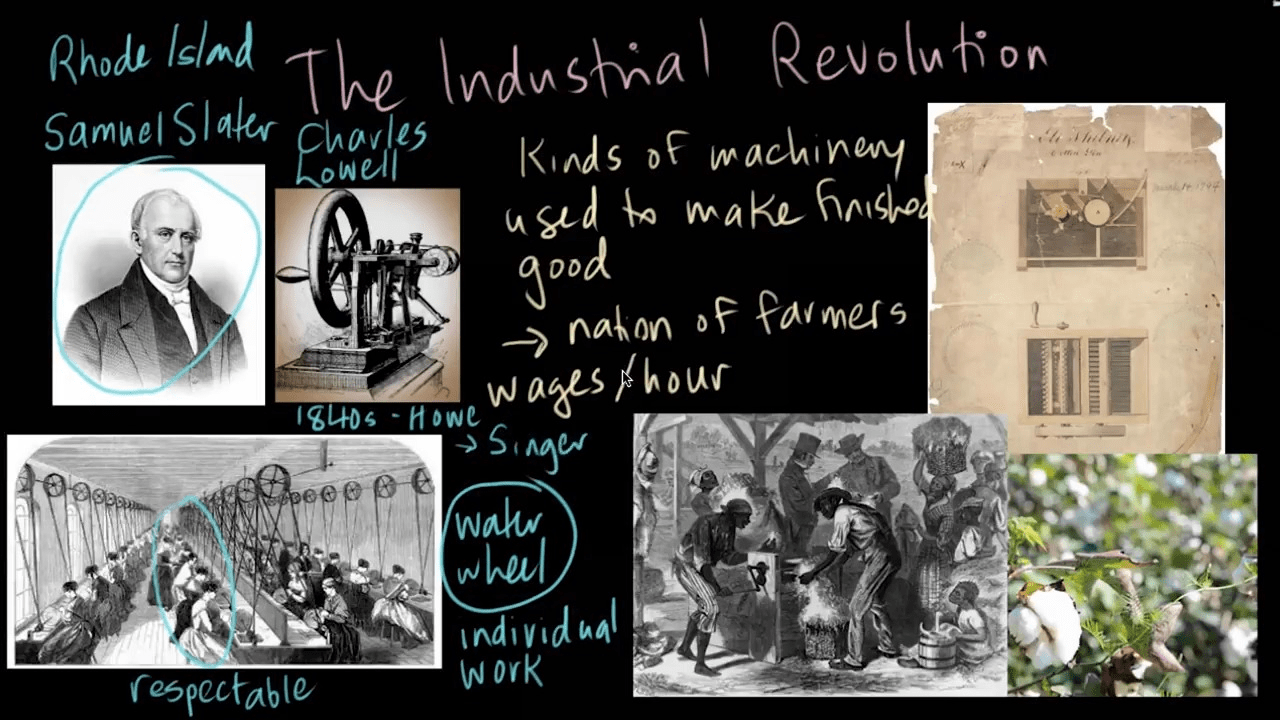}} \\
  Length & 12:58 & 21:55 \\
  Question 1 & What molecule, also called the biological blueprint for life, do all living organisms possess? & The activity shown in the image contributed most directly to which of the following? \\
  Question 2 & All living things on Earth share important processes. What are those four processes? & Which of the following developments most directly relates to the overall trend from 1800 to 1840 depicted on the graph? \\
  Question 3 & In 1828, a German chemist, Friedrich Wohler, proved that organic life is composed of the same components as inorganic matter. How did he do this? & According to the passage, which of the following best explains the most important effect that technological developments had on the American society? \\
  Question 4 & What ingredient(s) are considered necessary for the creation of life? & Which of the following had a significant long-term result of the major pattern depicted on the map? \\
  Question 5 & What quality of life allows for its slow diversification? & None \\
  \bottomrule
  \end{tabularx}
  \label{tab:video_info}
\end{table*}
    
\subsection{User Interaction}
FastPerson provides a personalized learning experience by allowing users to adjust the amount of time they spend watching a video and where they spend time. 
For example, users can quickly grasp the gist of an entire video by watching only the abridged version. 
Alternatively, users can view the abridged version for the parts they understand and the original detailed video for the parts they do not. 
The design of this learning experience is inspired by the flexibility of the information acquisition experience when reading a book. 
An example of how the system can be used to view a video is presented below.

\begin{enumerate}
  \item \textbf{Skimming through the entire summary}: By clicking the \textbf{Summary button} (C) and playing the video, users can watch the summarized version of the video. This is similar to the action of quickly skimming through the entire text content. 
  \item \textbf{Skimming over parts that are interesting}: Using the \textbf{Segments Container} (E), users can obtain a brief overview of each segment based on the associated thumbnail and text. Clicking on any segment will play its respective video after clicking the \textbf{Summary button} (C), enabling users to sample content before deciding to watch the detailed video. If the summary is sufficiently clear, users can move to the next segment, or if users want more details, they can check the original video of that segment for a better understanding by clicking the \textbf{Original button} (C). This is similar to the behavior of a user selecting a section of text that they like, quickly reading that section, and deciding whether the section really fits their interests.
  \item \textbf{Watching only the parts of interest repeatedly}: Users can replay any segment multiple times for a thorough understanding, which mirrors the experience of rereading one's favorite section of a book. 
  \item \textbf{Searching for specific content}: The \textbf{Search Container} (D) allows users to search for specific terms or keywords across segments, providing direct access to desired content, which is similar to using an index in a book.
  \item \textbf{Switching between the summary and the detailed view}: At any given point, users can toggle between the summarized and original videos using the \textbf{Summary and Original buttons} (C), catering to their comprehension and interest levels. This enables a behavior similar to that of a user reading text, where the parts of the text that the user fully understands or is interested in are read quickly, while the parts that the user does not fully understand or is interested in are read slowly.
\end{enumerate}

\section{Evaluation Experiments}

A series of quantitative and qualitative evaluation
experiments were conducted to assess the learning effectiveness and user experience of the FastPerson system,
The specific objectives were as follows:
\begin{itemize}
    \item To compare the learning effects when using the FastPerson system versus a standard video playback player.
    \item To evaluate the effectiveness of summarization using the FastPerson system and its feature of switching between the summarized and original videos in aiding understanding.
    \item To assess the usability and user satisfaction of the FastPerson interface.
\end{itemize}

\begin{figure*}[t]
  \centering
  \includegraphics[width=\linewidth]{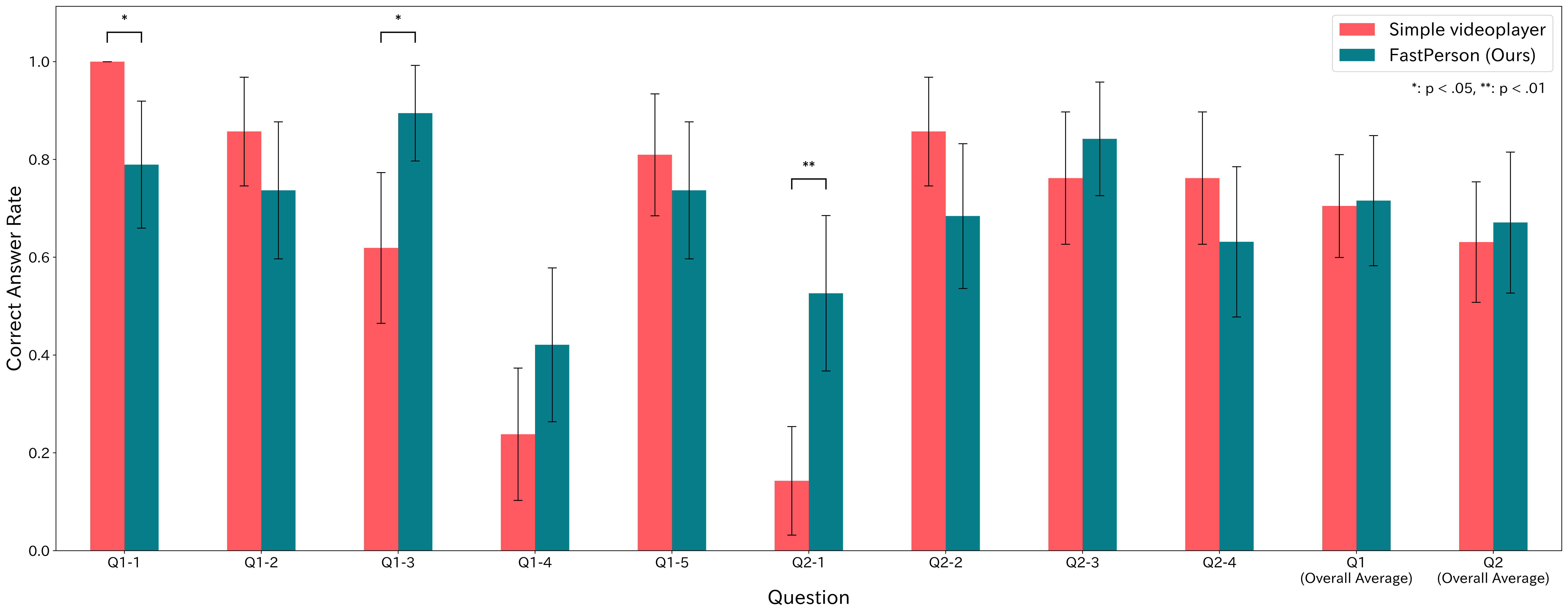}
  \caption{Correct Answer Rates for Each Question of the Video}
  \label{fig:exp1_1}
\end{figure*}

\subsection{Participants}
Participants from diverse backgrounds were recruited through the crowdsourcing platform Prolific.
A total of 40 participants participated in the study, randomly assigned to an intervention group that used FastPerson (19 participants) or the control group that used a generic video player (21 participants).
The participants were selected based on their fluency in English and completion of at least secondary eduction, because the videos were in English and the participants  needed to be able to understand the content of the videos.
The participants' demographics (gender, age, educational background, etc.) were collected as additional information to assess the effect of the experiment.
The participants comprised 23 males and 17 females, with an average age of 26.05 years ($\sigma=7.05$). 
Among the participants, 35, 35, 15, and 15\% had a high school diploma or GED, university bachelor's degree (BA/BSc/other), technical school or community college, and graduate degree (MA/MSc/MPhil/etc.), respectively.
A post experiment questionnaire measured the participants' prior knowledge of the videos they watched on a 5-point scale (1 being ``no prior knowledge'' and 5 being ``extensive knowledge''), and the results showed that, for the first video, the intervention and control groups had means of 2.63 ($\sigma=1.16$) and 2.10 ($\sigma=0.94$), respectively. For the second video, the intervention and control groups had means of 1.32 ($\sigma=0.58$) and 1.38 ($\sigma=0.59$), respectively.
A t-test between the two groups revealed t-statistic values of 1.607 with a p-value of 0.116 and -0.351 with a p-value of 0.727 for the first and second videos, with no statistically significant difference between the two groups at the 0.05 level of significance.
Thus, there was no clear difference in preconditioning knowledge between the two groups. 
Further, the participants were motivated to achieve high scores because it was clearly stated that the experiment participants with the highest scores would receive a bonus reward.
All the participants in the experiment understood and agreed that the process was voluntary and that they had the right to withdraw at any time.

\subsection{Procedure}
In this experiment, participants gathered through crowdsourcing were randomly assigned to an intervention group using FastPerson and a control group that used a general video player. 
Two videos were selected from Khan Academy, an online learning service, for viewing, and the accompanying questions were used to measure the learning effect of the videos. 
To prevent the participants' understanding of the videos from changing drastically based on their prerequisite knowledge, we selected lecture videos with a level of difficulty that could be understood by the video alone.
The selected videos were from the fields of biology and history, as shown in Table~\ref{tab:video_info}. 
The videos were accompanied by a number of questions to judge the participants' understanding of its content.
Several questions for the first video could not be answered  correctly without looking at materials besides the video, and therefore, five questions excluding these ones were used to check the participants' understanding of the video. 
For the second video, the four questions included in the video were used to assess comprehension. 
The first video is a lecture video in which the focus is on the speaker and the visual elements, while in the second video, the focus is on the board. 
The participants viewed these videos in the manner assigned to their confidence, and subsequently, they were asked to answer similar questions. 
For the intervention group that used FastPerson, an additional questionnaire about the system was collected. The experiment was conducted through Prolific, with a median working time of 46 min and 26 s, and each participant was paid 7\pounds.

\begin{figure}[t]
  \centering
  \includegraphics[width=0.9\linewidth]{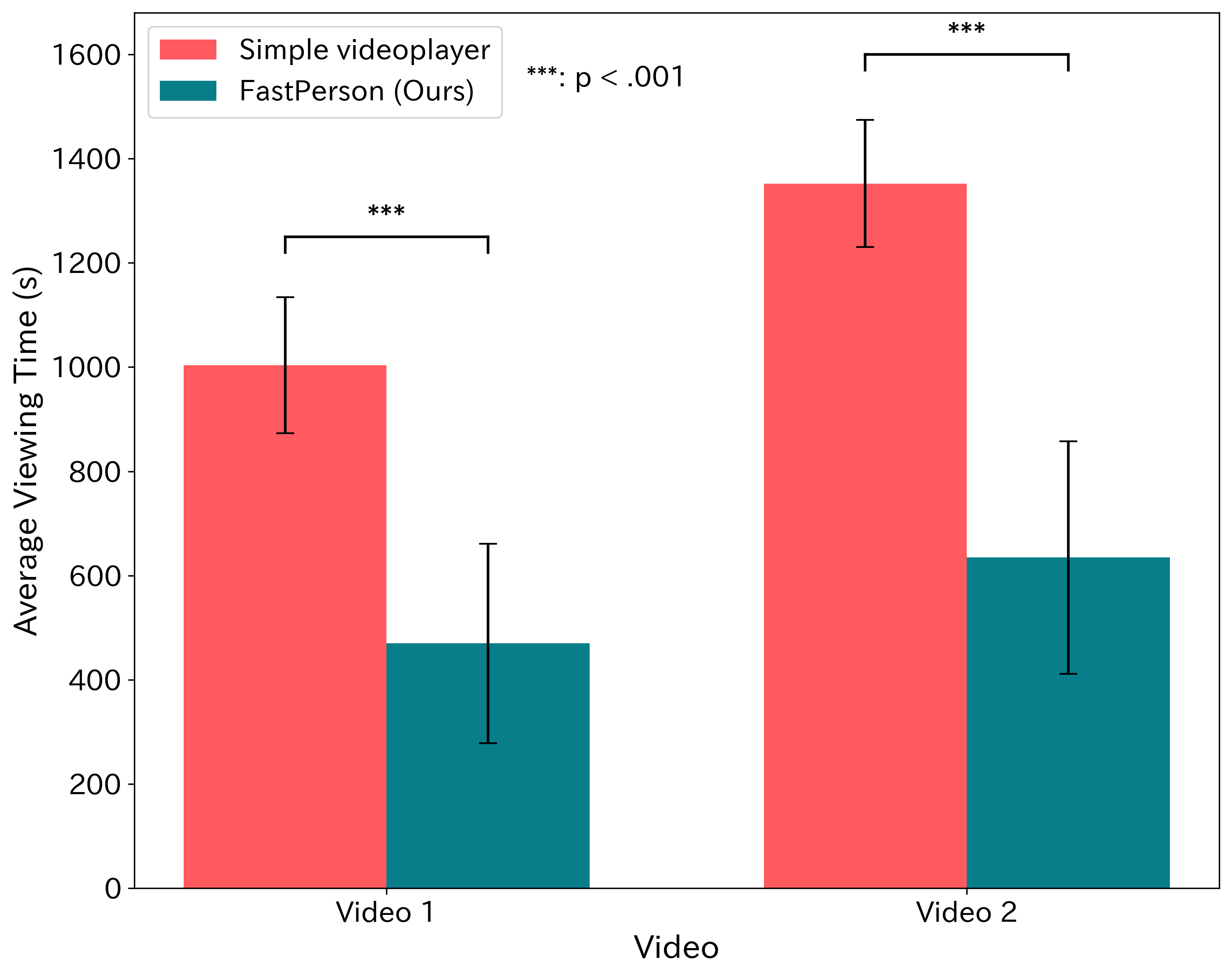}
  \caption{Average Video Viewing Time by Viewing Method}
  \label{fig:exp1_2}
\end{figure}

\subsection{Evaluation of Learning Effectiveness}
We analyzed the percentage of correct answers to questions about the videos and the time taken to view the videos after watching the same two videos for the control and the intervention groups to evaluate the learning effectiveness.
The right side of Fig.~\ref{fig:exp1_1} shows graphs comparing the average correct response rate of users to the questions in Videos 1 and 2 (``Q1'' and ``Q2,'' respectively).
For Video 1, the correct answer rate for the control group (red bar) was 0.70 ($\sigma=$ 0.33) and that of intervention group (blue bar) was 0.71 ($\sigma=$ 0.42) while those for Video 2 are 0.63 ($\sigma=$ 0.39) and 0.67 ($\sigma=$ 0.46), respectively. 
Therefore, there is no clear difference in the percentage of correct answers between both groups.
In fact, a t-test between the two groups yields t-statistic values of 0.170 with a p-value of 0.864 and 0.528 with a p-value of 0.598 for the first and second videos, respectively, neither of which is statistically significant when the significance level is set at 0.05.
Further, the percentage of correct answers for each question (left side of Fig.~\ref{fig:exp1_1}) shows that the score for viewing on a general video player was higher for question 1 of video 1 (Q1-1), while the score for viewing on FastPerson was higher for question 3 of video 1 (Q1-3) and question 1 of video 2 (Q2-1).
Q1-1 is ``What molecule, also called the biological blueprint for life, is found in all living organisms?'' and the correct answer is ``DNA,'' which is picked up as a keyword in the FastPerson summary, indicating that it is a keyword that ``all living organisms possess.'' 
However, the percentage of correct answers is likely to have decreased because of a lack of explanation of the term ``biological blueprint for life.''
Meanwhile, although Q1-3 and Q2-1 are adequately covered in both the simple video playback and FastPerson, the topics of the other incorrect choices in the question are relatively fewer in FastPerson. 
Therefore, users who watched FastPerson are not expected to select these incorrect choices, resulting in a higher percentage of correct answers.
The time taken to watch the videos is summarized in Fig.~\ref{fig:exp1_2}, where the average time taken by users to watch Videos 1 and 2 was 1003 ($\sigma=312$) and 1352 ($\sigma=270$) s, respectively, using the regular video player. 
FastPerson, which provides summary videos, took users an average of 469 ($\sigma=424$) and 634 ($\sigma=496$) s to watch Videos 1 and 2, respectively.
Using FastPerson reduced the viewing times for Videos 1 and 2 by 53.24 and 53.11\%, respectively.
A t-test between the two groups showed that the t-statistic values for the first and second videos were 4.409, with a p-value of $1.063\cdot10^{-3}$, and 5.281, with a p-value of $9.955\cdot10^{-6}$, respectively, both of which are statistically significant differences at $p<0.001$. 
Therefore, learning with FastPerson enables us to reach a level of understanding identical to that of the original video in a shorter time compared to that when watching the original video.

\subsection{Evaluation of User Experience}
\begin{figure*}[t]
  \centering
  \includegraphics[width=\linewidth]{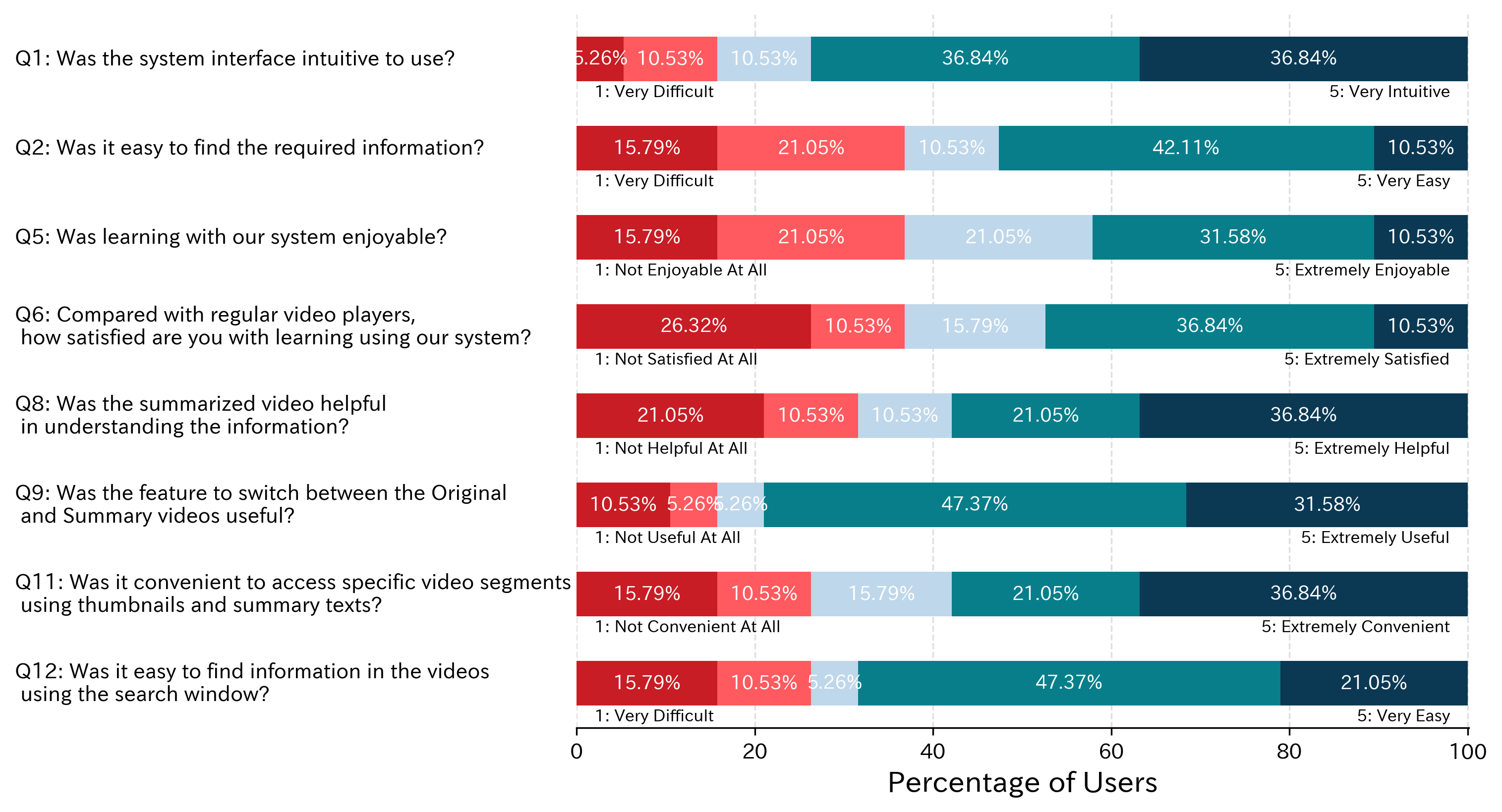}
  \caption{Survey Results on User Experience}
  \label{fig:exp2_1}
\end{figure*}

We conducted a questionnaire survey on the proposed method after 19 participants in the FastPerson intervention group finished learning from the videos to evaluate the usability, learning effectiveness, and user experience of the system.
The survey included the following aspects:
\begin{enumerate}
  \item Interface Usability
  \begin{itemize}
    \item[Q1:] Was the system interface intuitive to use? (1: Very Difficult - 5: Very Intuitive)
    \item[Q2:] Was it easy to find the required information? (1: Very Difficult - 5: Very Easy)
    \item[Q3:] How was your experience of using the Summary and Original buttons? (Open-ended Response)
    \item[Q4:] What improvements would you suggest for the user interface design? (Open-ended Response)
  \end{itemize}
  \item Learning Experience Satisfaction
  \begin{itemize}
    \item[Q5:] Was learning with our system enjoyable? (1: Not Enjoyable At All - 5: Extremely Enjoyable)
    \item[Q6:] Compared with regular video players, how satisfied are you with learning using our system? (1: Not Satisfied At All - 5: Extremely Satisfied)
    \item[Q7:] How would you rate your learning experience with our system compared to that with other methods? (Open-ended Response)
  \end{itemize}
  \item Usefulness of Summary and Original Videos
  \begin{itemize}
    \item[Q8:] Was the summarized video helpful in understanding the information? (1: Not Helpful At All - 5: Extremely Helpful)
    \item[Q9:] Was the feature to switch between the Original and Summary videos useful? (1: Not Useful At All - 5: Extremely Useful)
    \item[Q10:] How did you perceive the quality of the summarized videos? (Open-ended Response)
  \end{itemize}
  \item Evaluation of Specific Features
  \begin{itemize}
    \item[Q11:] Was it convenient to access specific video segments using thumbnails and summary texts? (1: Not Convenient At All - 5: Extremely Convenient)
    \item[Q12:] Was it easy to find information in the videos using the search window? (1: Very Difficult - 5: Very Easy)
  \end{itemize}
\end{enumerate}
The questions included both numerical and open-ended questions, and the results for the numerical questions are shown in Fig.~\ref{fig:exp2_1}.
For each question, the percentage of users who responded to each of the five levels is presented as a bar graph.

\subsubsection{Evaluation of Interface Usability}
The average rating was 3.89 ($\sigma=1.20$) for the question about the system's ease of use, especially intuitiveness (Q1).
This indicates that, on average, the users found the interface intuitive.
For the question about the ease of finding information in the system (Q2), the average rating was 3.11 ($\sigma=1.33$).
This rating reflects that users find it moderately easy to find the required information.
However, the relatively high standard deviation highlights the diversity in user experience, indicating that some users were able to navigate and find information more easily than the others.
The open-ended questions provided additional insights.
For example, in terms of the usability of the Summary and Original buttons, many participants states that they were ``very easy to use and understand'' and they appreciated the usefulness and convenience of switching between the different video modes (Q3).
However, one user commented, ``The interface is all over the place, and you have to spend a lot of time to understand it.'' 
This suggests that there is room for improvement in the placement of buttons and components.
Suggestions for interface design improvements received a wide range of responses (Q4).
Although some participants felt that the humanization of the summary descriptions and voice reading needed further improvements, others called for improvements to the overall visual aspect of the site.
General suggestions included improving the monotony of the design and increasing the visibility of the keyword search function.
These suggestions underscored the desire for a visually appealing and functionally robust interface.
Another comment was that ``the videos are fragmented, so users have to wait for each part to finish before they can move on to another part.''
In fact, the users could move freely to another part by clicking on the thumbnail of the video segment; however, this was not clearly communicated to the user, indicating room for improvement.

\subsubsection{Evaluation of Learning Experience Satisfaction}
A quantitative measure of the enjoyment of learning with FastPerson was obtained with a mean rating of 3.00 (Q5).
Although this rating indicates a moderate level of enjoyment among users, the standard deviation of 1.29 suggests that there is a wide range of variation in the participants' experiences.
Next, we measured learning satisfaction compared to that with the regular video player and found a slightly lower mean satisfaction of 2.95 ($\sigma=1.43$) (Q6).
This could be attributed to some users having a preference or comfort level with using traditional video players.
Higher standard deviations indicate larger differences in user satisfaction, suggesting that they may not perfectly match the learning preferences and expectations of all users.
In the qualitative feedback received through the open-ended question Q7, users who rated their experience positively indicated that the features of FastPerson that allowed them to quickly obtain the information they needed and the interactive nature of the learning experience contributed to their enjoyment and satisfaction.
In contrast, users who rated their experience as unsatisfactory cited unfamiliarity with the interface and a preference for the simpler approach of traditional video players, suggesting that a simpler presentation of the components could contribute to user satisfaction.

\subsubsection{Evaluation of the Usefulness of Summary Videos}
The mean rating for the usefulness of the summary videos in understanding the information was 3.42 ($\sigma=1.61$) (Q8).
This rating indicates a moderately positive acceptance of the effectiveness of the summary videos in conveying information.
Similarly, the average rating for the ability to switch between summary and original video formats was 3.84 ($\sigma=1.26$) (Q9).
Users appreciated the ability to switch between in-depth and summary video formats, giving them more control over the learning process.
In the qualitative feedback (Q10), many users praised the concept of the summary videos, noting that they could help them quickly understand key concepts, especially when reviewing a topic or when time is limited.
However, they raised concerns about the quality of the content in the summaries.
The main issues were related to the depth of content and the narration of the summary video.
The summaries were found to be useful; however, they sometimes lacked sufficient detail to fully understand complex topics.
In addition, several users pointed out that the pronunciation of the words in the summary video was different, confirming that the quality of the summary video could be improved by using a more powerful speech synthesizer.

\subsubsection{Evaluation of Specific Features}
The ease of accessing video segments using thumbnails and summary text received an average rating of 3.53 ($\sigma=$1.50) (Q11).
Similarly, the ease of retrieving information in the video using the search window received an average rating of 3.47 ($\sigma=$1.39) (Q12).
These features are not unique to FastPerson; however, they are compatible with the ability to switch between the summary and original videos.
The majority of users rated both the thumbnail and summary text as well as the ability to switch between the original and summarized videos (average rating 3.84) as useful, which indicates that presenting the user with both the summary video and summary text, rather than just the summary text of the video, improves usability and convenience.

\section{Discussion}
\subsection{Advantages of FastPerson in Educational Content Summarization}
FastPerson has shown significant advantages in learning through the use of educational video content. 
It can efficiently reduce video viewing time by 53\% and achieve the same level of comprehension as with the original video. 
Interestingly, for questions related to important parts of the video, FastPerson occasionally produces a higher percentage of correct answers than that with the original full-length video, suggesting that it can effectively focus on the important parts of the content. 
The ability to switch between summary and original video formats was particularly appreciated by users, offering flexibility in learning and content consumption that is not available in traditional methods. 
This feature allows learners to quickly grasp essential concepts through summaries and delve into detailed explanations as needed, thereby catering to diverse learning preferences and needs. 
In addition, FastPerson's use of visual elements synthesized with summarized audio enhances the educational value of the content, making complex information more accessible and engaging, especially for visual learners.

\subsection{Identified Limitations and Areas for Improvement}
FastPerson has some limitations. 
For some questions, the percentage of correct answers was higher when the original video was viewed, suggesting that information may be missed in the summarized video. 
We believe that the system can be improved by encouraging users to view the original video when information is missing. 
Further, users have also indicated that transitions between video segments are not seamless and that the sound quality needs improvement.
We anticipate that these issues could be addressed by improving the implementation and replacing it with a higher quality speech synthesis method.
Other users also pointed out the need for a more intuitive user interface because they found it difficult to understand the functionality of the various features. 
In addition, we observed that some users prefer the original simple video playback player, and we believe that simplifying the overall structure of FastPerson can help improve user accessibility. 

\subsection{Technological Enhancements and Future Directions}
FastPerson extracts visual elements to be synthesized with the summarized audio from the original video. 
Using a text-to-video model, other methods can be employed to generate a video from the summarized text; Runway's Gen-2 and OpenAI's Sora are possible candidates for these methods. 
These methods focus on generating object-centered videos, and therefore, it currently requires more work to create lecture videos that are text-heavy and slide-centered. 
However, it may be possible to create helpful summary videos for generating lecture videos focused on some topics. 
For example, if a lecture is centered around the intricacies of plant biology, a text-to-video model could be used to generate video segments illustrating key concepts such as photosynthesis, cellular structures of plants, or growth cycles. 
This approach would not only make the content more accessible but also more engaging for viewers, especially for visual learners.

\subsection{Broader Implications and Applications of FastPerson}
The broader implications of FastPerson go beyond its immediate application in educational video summarization. 
FastPerson has the potential to revolutionize how we interact with a wide range of video content. 
For example, in professional settings, FastPerson could be used to distill long training sessions or meetings into concise summaries, allowing for efficient review and better time management. 
Further, FastPerson has potential applications related to media consumption, where users can quickly view news or entertainment highlights. 
In addition, its application could extend to content creators and marketers, giving them a video editing tool to create engaging, summarized versions of their content for better audience engagement. 
FastPerson's adaptability could enable its integration into various platforms, making information more accessible and digestible for a broader audience.

\section{Conclusion}
In this study, we proposed a new video summarization technology called "FastPerson" that comprehensively considers visual and auditory information in lecture videos to improve the efficiency of the learning experience. 
FastPerson produces a summary video by converting visual and auditory elements of a video into text and generating a summary sentence that integrates them. 
This system helps learners understand important information quickly and efficiently, especially when they have limited study time or are interested in multiple topics. 
The results of the evaluation experiments indicated no significant difference in video comprehension when using FastPerson and the conventional video playback methods; however, the former successfully reduced viewing time by ~53\%.
This result suggests that FastPerson can provide an efficient learning experience. 
In addition, the usability and interaction feedback demonstrates the intuitiveness of the interface and usefulness of the ability to switch between summary and original videos. 
The results also show room for improvement in terms of the depth of the summary content and the quality of the audio.

\begin{acks}
  This work was supported by JST Moonshot R\&D Grant JPMJMS2012, JST CREST Grant JPMJCR17A3, and the commissioned research by NICT Japan Grant JPJ012368C02901.
\end{acks}

\bibliographystyle{ACM-Reference-Format}
\bibliography{reference}

\end{document}